\documentclass[5p,times]{elsarticle}

\usepackage{lineno,hyperref}
\usepackage{amssymb}
\usepackage{amsmath}
\usepackage{color}
\usepackage{hyperref}

\usepackage[figuresright]{rotating}

\begin{document}

\begin{frontmatter}

\title{A self-organizing neural network architecture for learning \\human-object interactions}

\author{Luiza Mici\corref{mycorrespondingauthor}}
\cortext[mycorrespondingauthor]{Corresponding author}
\ead{mici@informatik.uni-hamburg.de}

\author{German I. Parisi\corref{nocorrespondingauthor}}
\ead{parisi@informatik.uni-hamburg.de}

\author{Stefan Wermter\corref{nocorrespondingauthor}}
\ead{wermter@informatik.uni-hamburg.de}

\address{Knowledge Technology, Department of Informatics, University of Hamburg, Germany}

\begin{abstract}
The visual recognition of transitive actions comprising human-object interactions is a key component for artificial systems operating in natural environments.
This challenging task requires jointly the recognition of articulated body actions as well as the extraction of semantic elements from the scene such as the identity of the manipulated objects. 
In this paper, we present a self-organizing neural network for the recognition of human-object interactions from RGB-D videos. 
Our model consists of a hierarchy of Grow-When-Required (GWR) networks that learn prototypical representations of body motion patterns and objects, accounting for the development of action-object mappings in an unsupervised fashion.
We report experimental results on a dataset of daily activities collected for the purpose of this study as well as on a publicly available benchmark dataset.
In line with neurophysiological studies, our self-organizing architecture exhibits higher neural activation for congruent  action-object pairs learned during training sessions with respect to synthetically created incongruent ones.
We show that our unsupervised model shows competitive classification results on the benchmark dataset with respect to strictly supervised approaches.
\end{abstract}


\end{frontmatter}

\section{Introduction}

The recognition of transitive actions, i.e., actions that involve the interaction with an object, represents a key function of the human visual system for goal inference and social communication.
The study of transitive actions such as grasping and holding has often been the focus of research in neuroscience and psychology~\cite{Fleischer,nelissen2006charting}.
Nevertheless, this task has remained an open challenge for computational models of action recognition.

The ability of computational approaches to reliably recognize human-object interactions can establish an effective cooperation between assistive systems and people in real-world scenarios, promoting learning from demonstration in robotic systems~\cite{prevete2008connectionist,tessitore2010motor}.
Given the outstanding capability of humans to infer the goal of actions from the interaction with objects, the biological visual system represents a source of inspiration for developing computational models.
From the computational perspective, an important question arises regarding the potential links between the representations of body postures and manipulated objects and, in particular, how these two representations interact and can be integrated.

In the visual system, the information about body pose and objects are processed separately and reside in distinct subcortical areas~\cite{beauchamp2002parallel,downing2011role,grill2013representation}.
Neuroscientists have widely studied object and action perception, with a focus on where and how the visual cortex constructs invariant object representations~\citep{HubelAndWiesel} and how neurons in the superior temporal sulcus (STS) area encode actions in terms of patterns of body posture and motion~\cite{grossman2002brain,giese2003neural}.
It has been shown that the identity of the objects plays a crucial role for the complete understanding of human-object interactions~\citep{saxe2004understanding} and modulates the response of specific action-selective neurons~\cite{gallese1996action, nelissen2005observing,yoon2012neural}.
Yet, little is known about the exact neural mechanisms underlying the integration of actions and objects.

In this paper, we present a self-organizing neural architecture that learns to recognize human-object interactions from RGB-D videos.
The design of the proposed architecture relies on the following assumptions: (i) the visual features of body pose and man-made objects are represented in two distinct areas of the brain~\cite{downing2011role, grill2013representation, beauchamp2002parallel}, (ii) input-driven self-organization defines the topological structure of specific visual areas in brain~\citep{miikkulainen2006computational}, (iii) the representation of objects and concepts is based on prototypical examples~\citep{rosch1975family}, and (iv) the identity of the objects is crucial for the understanding of actions performed by other individuals~\cite{saxe2004understanding, gallese1996action}.

We develop a hierarchical architecture with the use of growing self-organizing networks, namely the Grow-When-Required (GWR) network~\citep{marsland2002self}, to learn prototypical representations of actions and objects and the resulting action-object mappings in an unsupervised fashion.
Growing self-organizing networks have been an effective model for clustering human motion patterns in terms of multi-dimensional flow vectors~\cite{parisiFrontiers, parisi2014human} as well as for learning object representations without supervision~\citep{donatti2010evolutionary}.
The generative properties of this topology of networks make them particularly suitable for our task when considering a possible generalization of unseen action-object pairs.

The proposed architecture consists of two network streams processing separately feature representations of body postures and manipulated objects.
A second layer, where the two streams are integrated, combines the information for the development of action--object mappings in a self-organized manner.
On the basis of previously reported results in Mici et al.~\citep{Mici2016}, this work contributes to improve the architecture design and provides a more in-depth analysis for an extended number of experiments.
Unlike our previous work, we use the GWR network for all layers including the object recognition module for which we employed a self-organizing map~(SOM)~\citep{Kohonen201352}.
The reason for this is the considerable impact on using a predefined topological structure~\citep{fritzke1993kohonen}, especially when having as input high-dimensional complex data distributions like perceptual representations of objects.
In our previous model, an additional network was used to learn prototypes of temporal activation trajectories of body poses before the integration phase.
However, the impact on the overall classification accuracy of the network was marginal, while it introduces more computational complexity.

We evaluate our architecture with a dataset of RGB-D videos containing daily actions acquired for the purpose of this study as well as with a publicly available action benchmark dataset CAD-120~\citep{koppula2013learning}.
We present and discuss our results on both datasets.
In particular, we look into  the role of the objects' identity as a contextual information for unambiguously distinguishing between different activities, the classification performance of our architecture in terms of recognition of human-object interaction activities, and the response of the network when fed with congruent and incongruent action-object pairs.
 
\section{Related work}\label{sec:material}

One important goal of human activity recognition in machine learning and computer vision is to automatically detect and analyze human activities from the information acquired from visual sensing devices such as RGB cameras and range sensors.
The literature suggests a conceptual categorization of human activities into four different levels depending on the complexity: gestures, actions, interactions, and group activities~\cite{aggarwal2011human,ziaeefard2015semantic,aggarwal2014human}.
Gestures are elementary movements of a person's body part and are the atomic components describing the meaningful motion of a person, e.g. \textit{stretching an arm} or \textit{raising a leg}.
Actions are single-person activities that may be composed of multiple gestures such as \textit{walking} and \textit{waving}.
Interactions are human activities that involve a person and one (or more) objects.
For instance, a \textit{a person making a phone call} is a human-object interaction.
Finally, group activities are the activities performed by groups composed of multiple persons or objects, e.g.~\textit{a group having a meeting}.

Understanding human-object interactions requires the integration of complex relationships between features of human body action and object identity.
From a computational perspective, it is not clear how to link architectures specialized in object recognition and motion recognition, e.g., how to bind different types of objects and hand/arm movements.
Recently,~Fleischer et al.~\citep{Fleischer} proposed a physiologically inspired model for the recognition of transitive hand-actions such as grasping, placing, and holding.
Nevertheless, this model works with visual data acquired in a constrained environment, i.e., videos showing a hand grasping balls of different sizes with a uniform background, with the role of the identity of the object in transitive action recognition being unclear.
Similar models have been tested in robotics, accomplishing the recognition of grip apertures, affordances, or hand action classification~\cite{prevete2008connectionist,tessitore2010motor}.

There is a number of techniques applied to the recognition of human-object interactions.
The most typical approaches are those that do not explicitly model the interplay between object recognition and body pose estimation~\cite{cippitelli2016human, yang2014effective, sung2012unstructured}.
Typically, first, objects are recognized and activities involving them are subsequently recognized, by analyzing the objects' motion trajectories~\citep{wu2007scalable}.
Yang et al.~\citep{yang2015robot} proposed a method for learning actions comprising object manipulation from demonstrating videos.
Their model is able to distinguish among different power and precision grasps as well as recognize objects by using a deep neural network architecture.
Nevertheless, the human action is simply inferred as the action with the maximum log-likelihood ratio computed over all possible trigrams \textless\textit{Object1}, \textit{Action}, \textit{Object2}\textgreater ~extracted from the sentences in the English Gigaword corpus.
Pieropan et al.~\cite{pieropan2014audio} proposed including action-related audio cues in addition to the spatial relation among objects in order to learn object manipulations for the purpose of robot learning by imitation. 
However, important descriptive visual features like body motion or fine-grained cues like the hand pose during manipulation were not considered.

Probabilistic approaches have been extensively used for reasoning upon relationships and dependencies among objects, motion, and human activities.
Gupta et al.~\cite{gupta2007objects, gupta2009} proposed a Bayesian network model for integrating the appearance of manipulated objects, human motion, and reactions of objects.
They estimate \textit{reach} and \textit{manipulation} motion by using hand trajectories as well as hidden Markov models (HMMs).
The Bayesian network integrates all of this information and makes a final decision to recognize objects and human activities.
Following a similar probabilistic integration approach,~Ryoo and Aggarwal~\cite{ryoo2007hierarchical} proposed a framework for the recognition of high-level activities.
They introduced an additional semantic layer providing feedback to the modules for object identification and motion estimation leading to an improvement of object recognition rates and better motion estimation.
Nevertheless, the subjects' articulated body pose was not considered as input data, leading to applications in a restricted task-specific domain such as airport video surveillance.
Other research studies have modeled the mutual context between objects and human pose through graphical models such as Conditional Random Fields (CRF)~\cite{yao2012recognizing, koppula2013learning, kjellstrom2011visual}. These types of models suffer from high computational complexity and require a fine-grained segmentation of the action sequences. 

Motivated by the fact that the visual recognition of complex human poses and the identification of objects in realistic scenes are extremely hard tasks, additional methods rely on extracting novel low-level visual features. 
Yao and Fei-Fei~\citep{yao2010grouplet} proposed a set of sophisticated visual features called \textit{Grouplet} which captures spatial organization of image patches encoded through SIFT descriptors \citep{lowe2004distinctive}. 
Their method is able to distinguish between interactions or just co-occurrences of humans and objects in an image, but no applications on video data have been reported.
Aksoy et al.~\citep{aksoy2011learning} proposed the \textit{semantic event chains} (SEC): a matrix whose entries represent the spatial relation between extracted image segments for every video frame.
Action classification is obtained in an unsupervised way through maximal similarity.
While this method is suitable for teaching object manipulation commands to robots, the representation of the visual stimuli does not allow for reasoning upon semantic aspects such as the congruence of the action being performed on a certain object.

Early attempts to apply neural networks for the problem of understanding human-object interactions from visual perception yielded promising results.  
Shimozaki and Kuniyoshi~\cite{shimozaki2003integration} proposed a SOM-based hierarchical architecture capable of integrating object categories, spatial relations, and movement and it was shown to perform well on simple 2D scenes of ball handling actions.
However, the literature suggests that compared to the static image domain, there is limited work on understanding human-object relationships from video data sequences with neural network architectures~\cite{lea2016segmental, 2017arXiv171106330M}. 

Systems for the estimation of articulated human body pose from 2D image sequences struggle through a great number of challenges such as changes in ambient illumination, occlusion of body parts and the enduring problem of segmentation.
The combination of RGB with depth information, provided by low-cost depth sensing devices such as Microsoft Kinect and Asus Xtion cameras, has shown computational efficiency in sensory data processing and has boosted a number of vision-based applications~\citep{HanMicrosoft}.
This sensor technology provides depth measurements which are used to obtain reliable estimations of 3D human body pose and tracking of body limbs in cluttered environments.
Applications of this type of technology have led to the successful classification of full-body actions and recognition of hand gestures~\citep{aggarwal2014human}.
However, a limitation of skeletal features is the lack of information about surrounding objects.
Wang et al.~\cite{Wang2014} proposed a new 3D appearance feature called \textit{local occupancy pattern}~(LOP) describing the depth appearance in the neighborhood of a 3D joint, and thus capturing the relations between the human body parts, e.g. hands, and the environmental objects that the person is interacting with.
Although their method produces state-of-the-art results, the identity of the objects is completely ignored, and the discriminative power of such features is unclear when the objects being manipulated are small or partially occluded.

\section{Methodology}\label{sec:arch}

\begin{figure*}
\begin{center}
\includegraphics[width=0.9\linewidth]{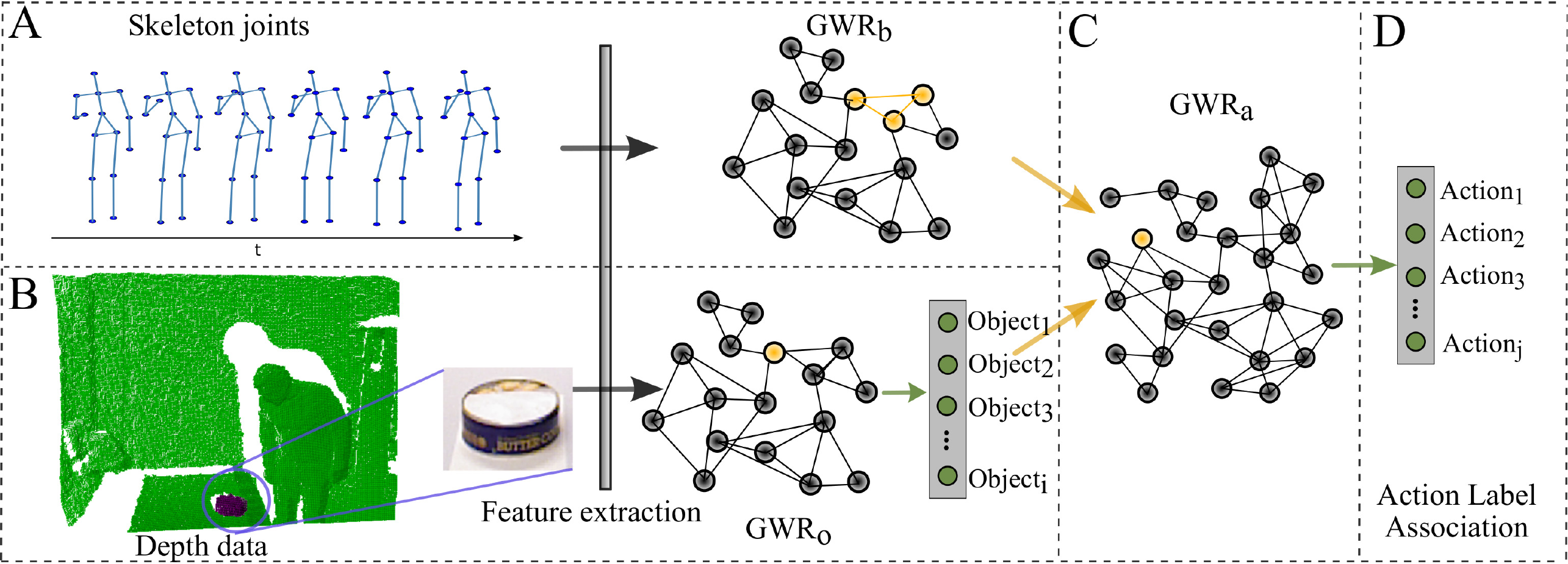}
\end{center}
\caption{Overview of the proposed architecture.~(A) Processing of the body postures. A set of local features that encode the posture of upper body limbs is extracted and fed to the \textit{GWR}$_b$ networks.~(B) The input for the
object recognition module is the RGB image of the manipulated object. The region of interest is automatically extracted through a point-cloud-based table-top segmentation. The object is represented as a compact feature vector and is fed to the \textit{GWR}$_o$ network which classifies the object. (C) The last network,~\textit{GWR}$_a$, learns the combinations of body postures and the object(s) involved in an action.~(D) Action labels are associated with each neuron in the \textit{GWR}$_a$ network in order to evaluate the architecture's action classification performance.}\label{fig:arch}
\end{figure*}

The proposed architecture consists of two main network streams processing separately visual representations of the body postures and of the manipulated objects. 
The information from the two streams is then combined for developing action-object mappings.
The building block of our architecture is the GWR network~\citep{marsland2002self}, which is a growing extension of the self-organizing networks with competitive learning.
An overview of the architecture is depicted in Fig.~\ref{fig:arch}.

The body pose cue is processed under the assumption that action-selective neurons are sensitive to the temporal order of prototypical patterns.
Therefore, the output of the body pose processing stream is computed by concatenating consecutively activated neurons of \textit{GWR}$_b$, with a sliding time window technique.
The object appearance cue is processed in order to have topological arrangements in \textit{GWR}$_o$ where different 2D views of 3D objects as well as different instances of the same object category are mapped to proximal neurons in the prototypes domain. 
The advantage of having such a topological arrangement consists in mapping any unseen view of a known object into the corresponding views learned during the training. 
This capability resembles, to some extent, biological mechanisms for learning the three-dimensional objects in the human brain~\citep{poggio1990network, perrett1996view, grill2013representation}.
Moreover, prototype-based learning approaches are supported by psychological studies claiming that semantic categories in the brain are represented by a set of most typical examples of these categories \citep{rosch1975family}.
For evaluating the architecture in terms of classification of human-object interaction activities, semantic labels are assigned to prototype neurons in \textit{GWR}$_a$ by extending the GWR algorithm with a labeling strategy.

\subsection{Learning with the GWR algorithm}\label{subsec:gwr}

Input-driven self-organization is an unsupervised mechanism that learns the input probability distribution through a finite set of prototype vectors.
Unlike traditional vector quantization (VQ) methods, self-organizing neural networks such as the SOM~\citep{Kohonen201352}, the neural gas (NG)~\citep{martinetz1991neural} as well as their growing extensions such as the growing neural gas (GNG)~\cite{fritzke1995growing} and the GWR algorithm~\cite{marsland2002self}, associate these prototype vectors with neurons that adaptively form topology preserving maps of the input space in an unsupervised fashion, i.e., similar inputs are mapped to neurons that are near to each other on the map.
This process of topology preservation is motivated by similar neural mechanisms found in multiple cortical areas of the brainx~\cite{miikkulainen2006computational}.

Growing self-organizing networks learn incrementally and can add (or remove) neurons according to different criteria.
Unlike the GNG algorithm, the neural growth of the GWR algorithm is not constant but rather depends on the overall network activation with respect to the input. 
This leads to a faster convergence and makes the GWR algorithm more suitable for learning representations of non-stationary datasets while being less susceptible to noise.

The GWR network is composed of a set of neurons associated with a weight vector and a set of edges that link the neurons forming neighborhood relationships.
The network starts with a set of two neurons randomly initialized and, during the learning iterations, both neurons and edges can be created, updated, or removed.
Given an input data sample $\textbf{x}(t)$, the index $b$ of the best-matching unit (BMU) is given by:

\begin{equation}\label{eq:bmu}
b = arg \min\limits_{j \in W} ||\textbf{x}(t) - \textbf{w}_j||,
\end{equation}
where $\textbf{w}_j$ is the weight vector of the neuron $j$ and $W$ is the set of all neurons.
The activity of the network $a$ is computed as a function of the Euclidean distance between the BMU, $\textbf{w}_b$, and the input data sample $\textbf{x}(t)$ at time step $t$:
\begin{equation}\label{eq:activation}
a = \exp ( { - ||\textbf{x}(t) - \textbf{w}_b|| } ).
\end{equation}
\noindent New neurons are added when the activity of the BMU is lower than an insertion threshold $a_T$.
This parameter modulates the amount of generalization, i.e., the largest discrepancy between an incoming stimulus and its BMU.
Edges are created between two neurons with the smallest distance from the input data sample, namely the first and the second BMUs.
Rarely activated edges (after $a_{max}$ iterations) and neurons without edges are removed.
Moreover, a firing rate mechanism that measures how often each neuron has been activated by the input leads to a sufficient training before new neurons are created. 
The firing rate is initially set to zero and than decreases every time a neuron and its neighbors are activated in the following way:
\begin{equation}\label{eq:hab}
\Delta h_i = \tau_i \cdot \kappa \cdot (1 - h_i) - \tau_i,
\end{equation}
where $\tau_i$, and $\kappa$ are the constants controlling the behaviour of the decreasing function curve of the firing counter.
Typically, the $\tau$ constant is set higher for the BMU ($\tau_b$) than for its topological neighbors ($\tau_i$).
Given an input data sample $\textbf{x}(t)$, if no new neurons are added, the weights of the winner neuron and its neighbors are updated as follows:
\begin{equation}\label{eq:update}
\Delta\textbf{w}_i = \epsilon_i \cdot h_i \cdot (\textbf{x}(t) - \textbf{w}_i),
\end{equation}

\noindent where $\epsilon_i$ and $h_i$ are the constant learning rate and the firing counter variable. 
The learning of the GWR algorithm stops when a given criterion is met, e.g., a maximum network size or a maximum number of learning epochs.

\subsection{Hierarchical learning}\label{subsec:data_proc}

We adopt hierarchical GWR learning~\cite{parisiFrontiers} for the data processing and subsequent action-object integration.
Hierarchical training is carried out layer-wise and in an offline manner with batch learning.
We first extract body pose, $A$, and object features, $O$, from the training image sequences, $T$ (see Section \ref{subsec:features}).
The obtained data is processed by training the first layer of the proposed architecture, i.e., \textit{GWR}$_b$ is trained with body pose data and \textit{GWR}$_o$ with objects~(Fig.~\ref{fig:arch}).
After training is completed, the \textit{GWR}$_b$ network will have created a set of neurons tuned to prototype body pose configurations, and the \textit{GWR}$_o$ network will have learned to classify objects appearing in each action sequence.    

\begin{figure}
\begin{center}
\includegraphics[scale=.6]{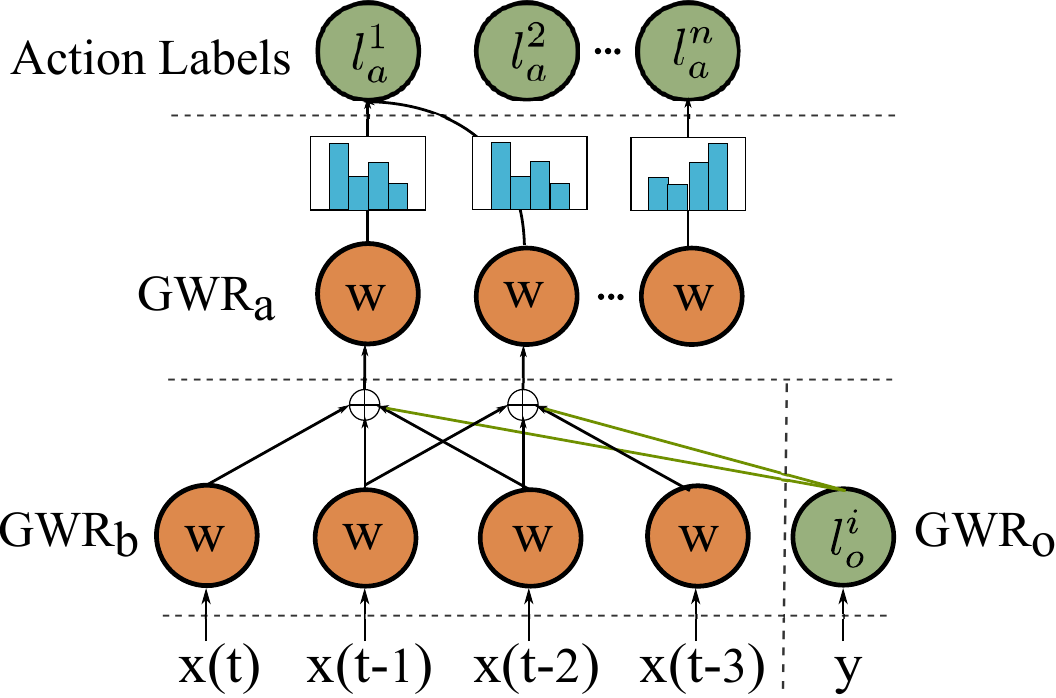}
\end{center}
\caption{Schematic description of the hierarchical learning and of the association of action labels (not all neurons and connections are shown). 
At each time step $t$, the input data sample $\textbf{x}(t)$ is represented by the weight $\textbf{w}$ of the winner neuron which is then concatenated with the previous winner neuron weights (two previous neurons in this example) and the category label of the object $l_o^i$ in order to compute the winner neuron in \textit{GWR}$_a$. 
Each \textit{GWR}$_a$ neuron is associated with a histogram of action categories, and the most frequently matched class will be the recognized action.}\label{fig:classification}
\end{figure}

The next step is to generate a new dataset $T^{*}$ for the \textit{GWR}$_a$ network that integrates information coming from both streams (Fig.~\ref{fig:classification}).
In order to encode spatiotemporal dependencies within the body pose prototypes space, we compute trajectories of the \textit{GWR}$_b$ best-matching units when having as input training action sequences.
For all body pose frames $\textbf{x}_i \in A$, the best-matching units are calculated as in Eq. \ref{eq:bmu} and the corresponding neuron weights are concatenated following a temporal sliding window technique, as follows: 
\begin{equation}\label{eq:trajectory}
\psi(\textbf{x}_i) = b(\textbf{x}_i) \oplus b(\textbf{x}_{i-1}) \oplus ... \oplus b(\textbf{x}_{i-q+1}), i \in [q,m],
\end{equation}
where $\oplus$ denotes the concatenation operation, $m$ is the total number of training frames, and $q$ is the width of the time window.
We will refer to the computed $\psi(\textbf{x}_i)$ by the name \textit{action segment}.

The object data $\textbf{y} \in O$ extracted from each action sequence is provided as input to the \textit{GWR}$_o$ network and the best-matching units $b(\textbf{y})$ are calculated as in~Eq.~\ref{eq:bmu}.
Objects are extracted only at the beginning of an action sequence. 
Therefore, the object representations to be learned contain no temporal information and the computation of neural activation trajectories, reported in Eq.~\ref{eq:trajectory}, is not performed. 
The label of the \textit{GWR}$_o$ best-matching unit is represented in the form of one-hot encoding, i.e., a vectorial representation in which all elements are zero except the ones with the index corresponding to the recognized objects' category.
When more than one object appears in one action sequence, the object data processing and classification with \textit{GWR}$_o$ is repeated as many times as the number of additional objects. The resulting one-hot-encoded labels are merged into one fixed dimension vector for the following integration step.

Finally, the new dataset $T^{*}$ is computed by concatenating each action segment $\psi(\textbf{x}_i)$ with the label of the corresponding object $l_o(\textbf{y})$ as follows:
\begin{equation}\label{eq:pair}
T^{*} = \{ \phi_u(\textbf{x}_i) \equiv \psi(\textbf{x}_i) \oplus l_o(\textbf{y}); \textbf{x}_i \in A, \textbf{y} \in O, u \in [q,m-q] \}.
\end{equation}
Each pair $\phi_u$, which we will refer to as an \textit{action-object} segment, encodes both temporally-ordered body pose sequences and the identity of the object being manipulated during the action sequence.
The \textit{GWR}$_a$ network is then trained with the newly computed dataset $T^{*}$, thereby learning the provided action-object pairs.

The resulting representative vectors of body pose can have a very high dimension which further increases when concatenating them through the temporal window technique. 
Methods based on the Euclidean distance metric, as in our case, are shown to have a performance degradation when data lies in high-dimensional space~\citep{aggarwal2001surprising}.
Therefore, we apply the principal component analysis (\textit{PCA}) dimensionality reduction technique to the neural weights of \textit{GWR}$_b$.
The number of principal components is chosen as a trade-off between accounting for the greatest variance in the set of weights and having a smaller dimensional discrepancy with the object's label.
The new basis is then used to project weights of activated neurons in \textit{GWR}$_b$ before the concatenation of the activation trajectories and the subsequent integration step.

\subsection{Classification}\label{subsec:classification}

We extend the GWR algorithm with a labeling strategy for classification tasks while keeping the learning process unsupervised.
We use a simple method based on the majority vote strategy as in \cite{strickert2005merge}. 
For each neuron $n_i$, we store information about the category of the data points it has matched during the training phase.
Thus, each neuron is associated with a histogram $hist(c, n_i)$ counting all cases of seeing a sequence with an assigned specific label $c$.
Additionally, the histograms are normalized by scaling the bins with the corresponding inverse class frequency $f_c$ and with the inverse neuron activation frequency $f_{a,n_i}$.
In this way, class labels that appear less during training are not penalized, and the vote of the neurons is weighed equally regardless of how often they have fired.
When the training phase is complete, each neuron that has fired during training, i.e., BMUs, will be associated with a histogram:
\begin{equation}
H_i = \frac{1}{f_c \cdot f_{a, n_i}} \cdot hist(c, n_i).
\end{equation}
At recognition time, given a test action sequence with length $k$, the best-matching units are computed for each frame and the action label $l$ is given by:
\begin{equation}\label{eq:label}
l = arg \operatornamewithlimits{max}\limits_{c} \left( \sum_{i=1}^{k} H_{b_i} \right).
\end{equation}
\noindent The classification of non-temporal data, e.g. object classification with the \textit{GWR}$_o$ network, is performed by applying majority vote only on the histogram associated to one best-matching unit $H_{bmu}$. 
This is a special case of Eq.~\ref{eq:label}, considering that $k=1$ for non-temporal data.

In our case, action sequences are composed of smaller action-object segments as described in Section~\ref{subsec:data_proc}.
Thus, the majority vote labeling technique described so far is applied as follows.
Let us assume we have a set of activity labels $L_a$ along with our training data, for instance, \textit{drinking} and \textit{eating}.
Therefore, each action-object segment $\phi \in T^*$ will be assigned with one of these labels and one action sequence will have the following form:
\begin{equation}
\Phi = \{ (\phi_1, l^j_a),...,(\phi_k, l^j_a), l^j_a \in L_a \},
\end{equation}
where $l^j_a$ is the activity label and $k$ is the number of action-object segments included in the sequence. 
During training of the \textit{GWR}$_a$ network on the action sequence $\Phi$, the label $l^a_j$ will be added to the histogram of the neurons activated for each of its composing segment $\phi$.
After the training is complete, the action sequence $\Phi$ will be classified according to the majority vote strategy (see Fig.~\ref{fig:classification}).
It should be noted that the association of neurons with symbolic labels does not affect the formation of topological arrangements in the network.
Therefore, our approach for the classification of objects and actions remains unsupervised.

\subsection{Feature extraction}\label{subsec:features}
\subsubsection{Body pose features}
Visual identification and segmentation of body pose from RGB videos are challenging due to the spatial transformations compromising the appearances, such as translations, the difference in the point of view, changes in ambient illumination, and occlusions.
For this reason, we use depth sensor technologies, such as the Asus Xtion camera, which provide us with reliable estimations of three-dimensional articulated body pose and motion even in real-world environments. 
Moreover, three-dimensional skeletal representations are the most straightforward way of achieving invariance to the subjects' appearance and body size.
We consider only the position of the upper body joints (\textit{shoulders}, \textit{elbows}, \textit{hands}, center of \textit{torso}, \textit{neck} and \textit{head}), given that they carry more significant information (than for instance the \textit{feet} and \textit{knee} joints) about the human-object interactions we focus on in this paper.
However, neither the number of considered joints nor the dimensionality of the input data limits the application of our architecture for the recognition of human-object interactions.

We extract the \textit{skeletal quad} features~\citep{evangelidis2014skeletal}, which are invariant with respect to location, viewpoint as well as body-orientation.
These features are built upon the concept of geometric hashing and have shown promising results for the recognition of actions and hand gestures. 
Given a quadruple of body joints $\{J_1, J_2, J_3, J_4\}$ where $J_i \in \mathbb{R}^3$, a local coordinate system is built by making $J_1$ the origin and mapping $J_2$ onto the vector $[1,1,1]^T$.
The position of the other two joints $J_3$ and $J_4$ are calculated with respect to the local coordinate system and are concatenated in a 6-dimensional vector $[\hat{j}_{3,1}, \hat{j}_{3,2}, \hat{j}_{3,3}, \hat{j}_{4,1}, \hat{j}_{4,2}, \hat{j}_{4,3}]$.
The latter becomes the compact representation of the four body joints' position.
We empirically select two quadruples of joints: [\emph{center torso, neck, left hand, left elbow}] and [\emph{center torso, neck, right hand, right elbow}].
This means that the positions of the hands and elbows are encoded with respect to the torso center and neck.
We choose the neck instead of the head position due to the noisy tracking of the head caused by occlusions during actions such as \textit{eating} and \textit{drinking}.

Composing such holistic body pose vectors, i.e., concatenations of joint positions, is quite convenient when employing a GWR network for the learning.
In the case of missing joints in a data frame, due to, for example, noise or body occlusion, the best-matching unit for that input vector can be computed omitting the missing parts of the body pose vector.
Self organizing networks, such as SOMs and the GWR networks as their growing extension, are able to operate robustly in the case of missing values~\cite{vatanen2015self}.

\subsubsection{Object features}

The natural variations in RGB images such as variations in size, rotation, and lighting conditions, are usually so wide that objects cannot be compared to each other simply based on the images' pixel intensities.
For this reason, we extract visual features from the object images in the following way. 
We extract dense SIFT features, which are not more than SIFT descriptors~\citep{lowe2004distinctive} computed at crossing points of fixed grids superimposed on each object image\footnote{Dense SIFT from VLFeat library:~\href{http://www.vlfeat.org/}{http://www.vlfeat.org/}}.
SIFT features have been successfully applied to the problem of unsupervised object classification \citep{tuytelaars2010unsupervised} and for learning approaches based on self-organization \citep{kinnunen2012unsupervised}.
Moreover, SIFT descriptors are known to be, to some extent, robust to changes in illumination and image distortion.
Multiple descriptors with four different window sizes are computed on every image in order to account for scale invariance between images.
The orientation of each of these descriptors is fixed and this relaxes the descriptors' invariance with respect to the object's rotation.
With this kind of representation, we can train the \textit{GWR}$_o$ network and obtain neurons tuned to different object views, yet invariant to translation and scale. 

\begin{figure}
\centering
\includegraphics[scale=.45]{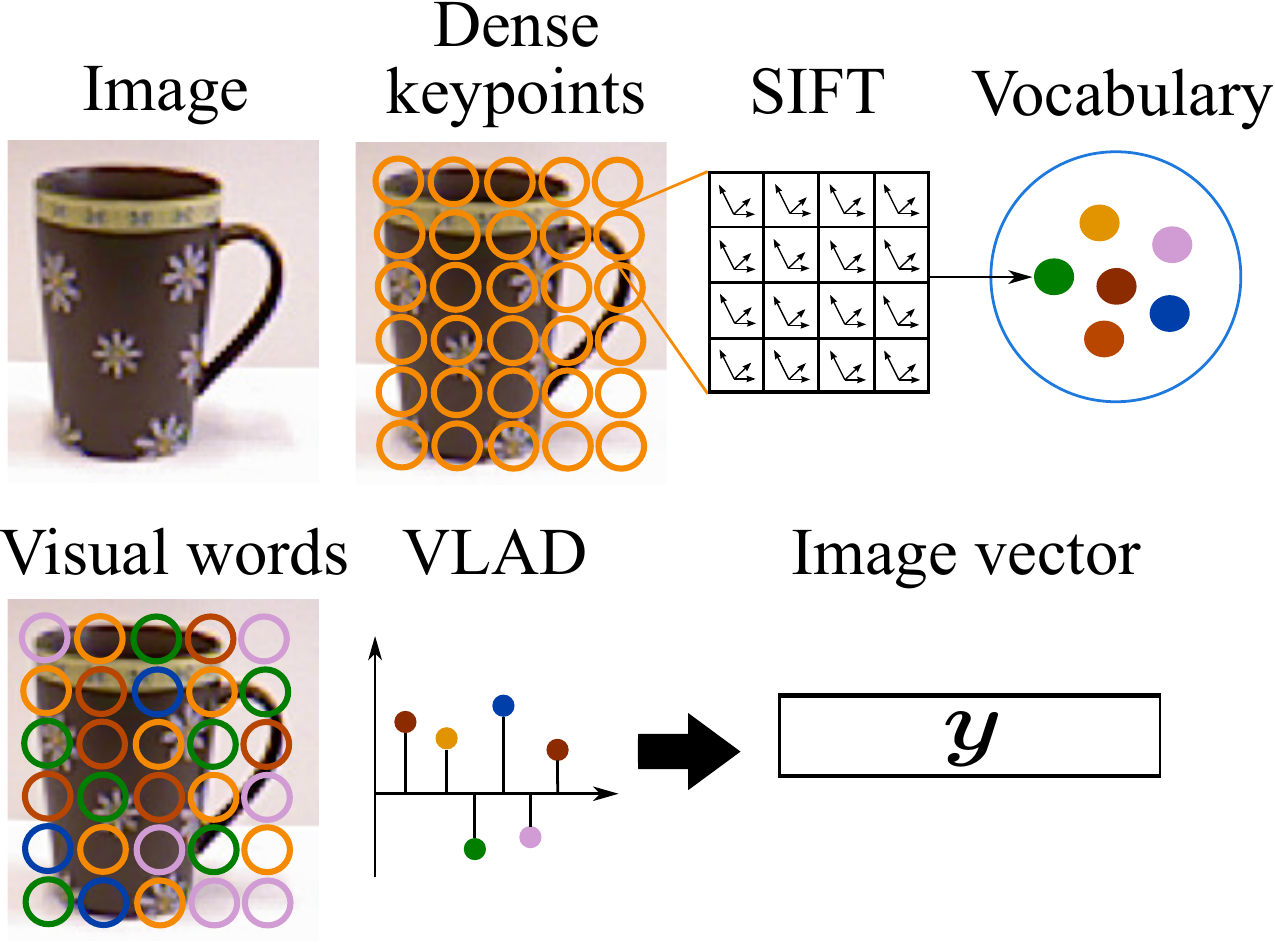}
\caption{Illustration of the steps for encoding object images with the VLAD encoding method.}
\label{fig:vlad}
\end{figure}

We perform quantization followed by an image encoding step in order to have a fixed-dimensional vectorial representation of each object image.
This is necessary since, during training of the \textit{GWR}$_o$ network, the objects are compared to each other through a vectorial metric, namely the Euclidean distance.
We apply the Vector of Locally Aggregated Descriptors~(VLAD)~\citep{jegou2012aggregating} encoding method (Fig.~\ref{fig:vlad}) which has shown higher discriminative power than the extensively used Bag of Visual Features~(BoF)~\cite{everingham2010pascal, szeliski2010computer}.
The BoF method simply computes a histogram of the local descriptors by hard assignment to a dictionary of visual words, whereas the VLAD method computes and traces the differences of all local descriptors assigned to each visual word.

\subsection{Training}\label{subsec:training}
In Table~\ref{table:parameters}, we report the parameters used for training the proposed neural architecture throughout the experiments presented in Section~\ref{sec:results}.
The selection of the range of parameters is made empirically
while also considering the GWR algorithm learning factors.
The parameters that we fix across all layers are the constants controlling the decrease function of the firing rate variable~($\tau_b$, $\tau_i$ and $\kappa$), the learning rates for the weights' update function ($\epsilon_b$ and $\epsilon_i$) and the threshold for the maximum age of the edges ($a_{max}$).
We set a higher insertion threshold parameter for the data processing layers, i.e., \textit{GWR}$_b$ and \textit{GWR}$_o$, than for the integration layer \textit{GWR}$_a$.
The higher value chosen for the \textit{GWR}$_b$ and \textit{GWR}$_o$ networks leads to a greater number of neurons created and a better representation of the input data as a result, whereas the slightly lower value for the \textit{GWR}$_a$ seeks to generate a set of neurons that tolerate more discrepancy in the input and generalize relatively more.
The insertion threshold parameters are very close to each other and very close to $1$, but their impact is not imperceptible given that the input data are normalized, i.e., take values within the interval $[0, 1]$.
We train each network for 300 epochs over the whole dataset in order to ensure convergence, during which the response of the networks to the input shows little to no significant modifications. 

\begin{table}
\caption{Training parameters of the \textit{GWR}$_b$, \textit{GWR}$_o$ and \textit{GWR}$_a$ networks of our architecture for the classification of human-object interactions.}
\label{table:parameters}
\footnotesize
\begin{center}
\begin{tabular}{l|c} 
Parameter & Value \\
\hline
Insertion Threshold & $a_T = \{0.98, 0.98, 0.9\}$\\
Firing Threshold & $f_T = 0.1$\\
Learning rates & $\epsilon_b = 0.1, \epsilon_i = 0.01$\\
Firing rate behavior & $\tau_b=0.3, \tau_i = 0.1,  \kappa = 1.05$\\
Maximum edge age &  $a_{max} = 100$\\
Training epochs & $300$ \\
\hline
\end{tabular}
\end{center}
\end{table}

In addition to the aforementioned parameters, the sliding window mechanism applied to processed body pose data also has an impact on the growth of the \textit{GWR}$_a$ network.
Wider windows lead to the creation of more neurons, albeit the slightly lower number of data samples.
This is an understandable consequence of the fact that the more temporal frames included in each time window, the higher the variance of the resulting data and the more prototype neurons created as a consequence.
However, this parameter has to be set empirically according to the experimental training data distribution.
We report the time window width parameter we set in each of our experiments in the following sections. 

\section{Experimental results}\label{sec:results}

We evaluated the proposed neural architecture both on the \textit{transitive actions dataset}~(Fig.~\ref{fig:mici_dataset}) that we have acquired for the purpose of this study and on a publicly available action benchmark dataset provided by the Cornell University, CAD-120~\cite{koppula2013learning}.
In this section, we provide details on both datasets, the classification performances obtained on these datasets, a quantitative evaluation of the integration module in the case of incongruent action-object pairs and a comparative evaluation on CAD-120.

\subsection{Experiments with the transitive actions dataset}
\subsubsection{Data collection}\label{subsection:collection}

\begin{figure}
\centering
\begin{minipage}[b]{.85\linewidth}
\centering\includegraphics[width=\linewidth]{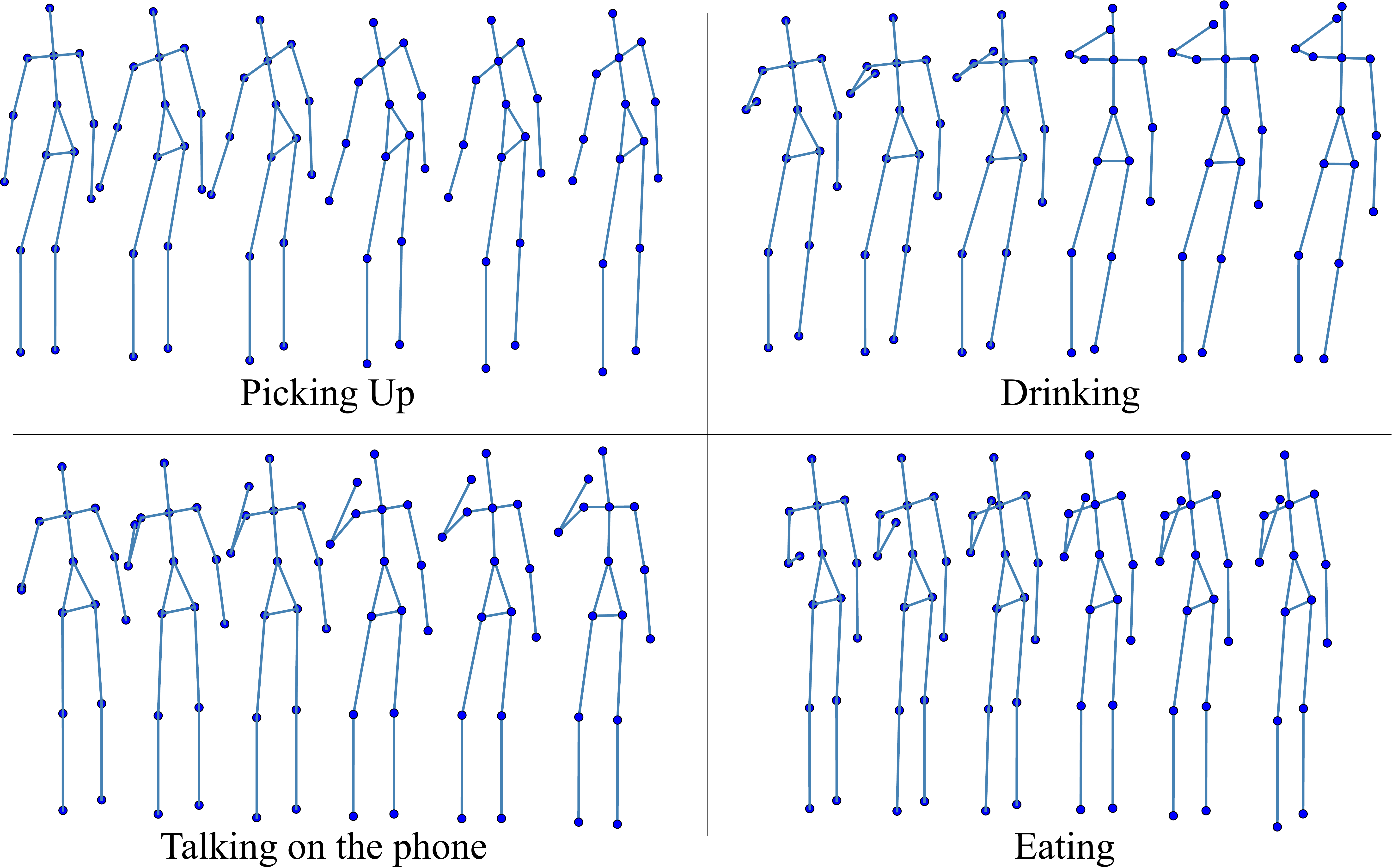}
\label{fig:dataset}
\end{minipage} \\
\begin{minipage}[b]{.6\linewidth}
\centering
\includegraphics[width=\linewidth]{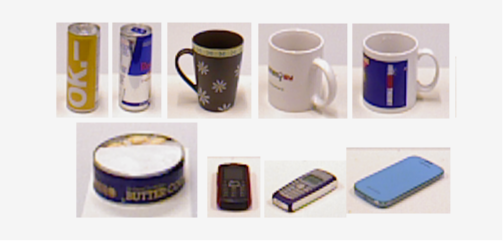}
\label{fig:mici_objects}
\end{minipage}
\caption{Examples of sequences of skeleton joints and objects taken from the transitive actions dataset. The object category labels are:~\textit{can}, \textit{mug}, \textit{biscuit box} and \textit{phone}.}\label{fig:mici_dataset}
\end{figure}

We collected a dataset of the following daily activities: \textit{picking up} (an object), \textit{drinking} (from a container like a mug or a can), \textit{eating} (an object like a cookie) and \textit{talking on the phone} (Fig.~\ref{fig:mici_dataset}).
The actions were performed by 6 participants that were given no explicit indication of the purpose of the study nor instructions on how to perform the actions.
The dataset was collected with an Asus Xtion depth sensor that provides synchronized RGB and depth frames at a frame rate of 30 fps.
The distance of each participant from the sensor was not fixed but maintained within the maximum range for the proper functioning of the depth sensor.
The tracking of the skeleton joints was provided by the OpenNI framework\footnote{OpenNI/NITE:~\href{
http://www.openni.org/software}{http://www.openni.org/software}}.
To attenuate noise, we computed the median value for each body joint every 3 frames resulting in 10 joint position vectors per second.
We added a mirrored version of all action samples to obtain invariance to actions performed with either the right or the left hand.
Action labels were then manually annotated.

The manipulated objects were segmented from each video using a point-cloud-based table-top segmentation algorithm\footnote{Point Cloud Library:~\href{http://www.pointclouds.org/}{http://www.pointclouds.org/}}, which extracts possible clusters on top of a plane surface, e.g., on the table. 
False positives obtained through the automatic segmentation were then manually deleted. 
Finally, the obtained images were used as training data for the object recognition module of our architecture. 

\subsubsection{Classification results}

We now assess the performance of the proposed neural architecture for the classification of the actions described in Section~\ref{subsection:collection}. 
In particular, we want to evaluate the importance of the identity of the manipulated object(s) in disambiguating the activity that a subject performs.
For this purpose, we conducted two separate experiments, whereby we process body pose cues alone and in combination with recognized objects.
Moreover, to further exclude any possible bias towards a particular subject, we followed a leave-one-subject-out strategy.
Therefore, six different trials were designed by using video sequences of the first five subjects for training and using the remaining subject for the testing phase. 
This type of cross-validation is quite challenging since different subjects perform the same action in a different manner and with a different velocity.

We trained each GWR network with the learning parameters reported in Section~\ref{subsec:training}. 
Since this dataset is composed of short temporal sequences, a time window of five frames was chosen for the concatenation of the processed body cues. 
This led to action-object segments of $0.5$ seconds, considering $10$ frames per second. 
When the training of the whole architecture was complete, the number of neurons reached for an input containing $\approx6500$ video frames was: $170$ neurons for the \textit{GWR}$_b$ network, $182$ for \textit{GWR}$_o$ and for the \textit{GWR}$_a$ network the number varied from $90$ to $120$ across different trials.

\begin{figure}
\begin{center}
\includegraphics[width=.65\linewidth]{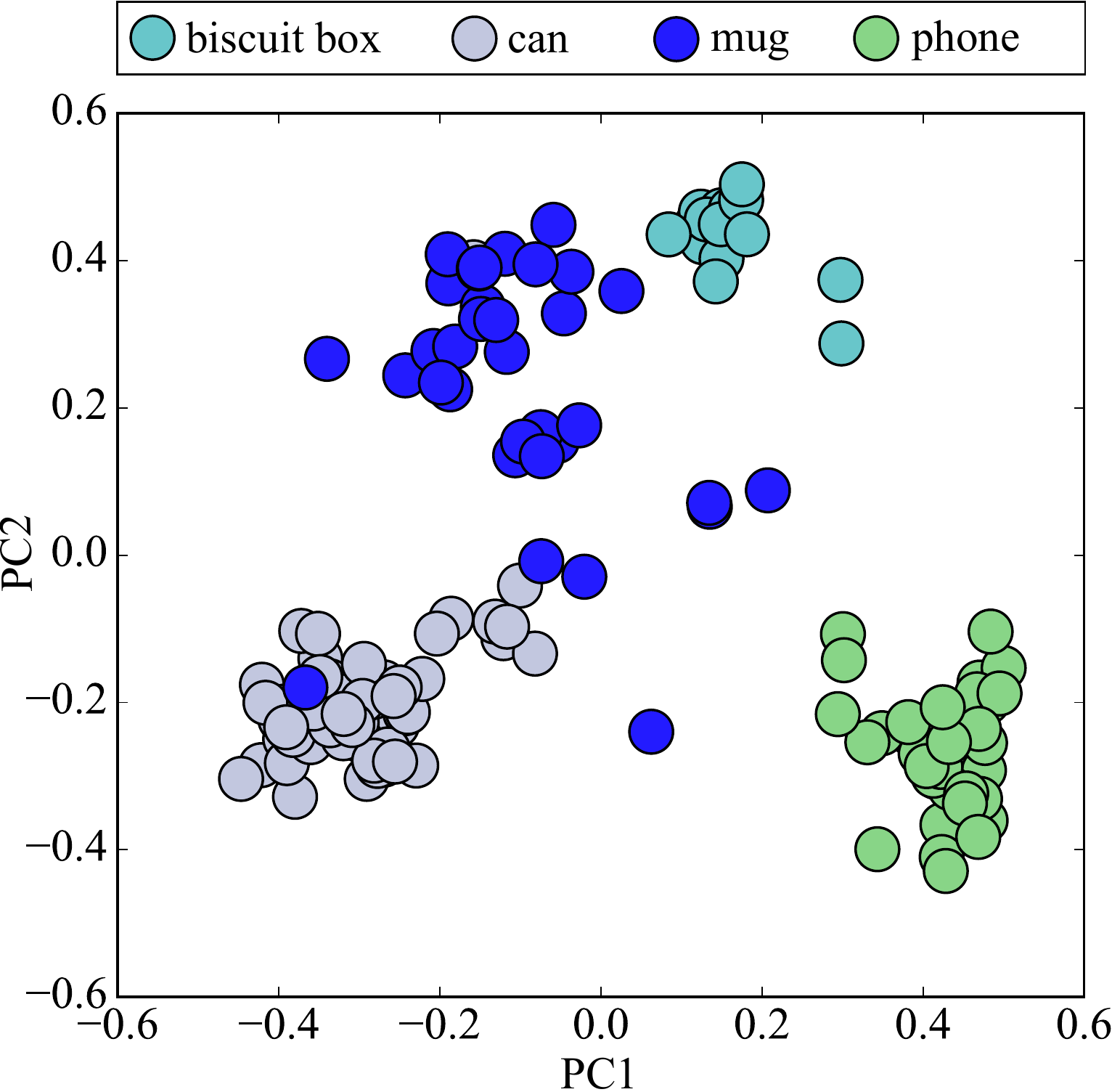}
\caption{Neural weights of the \textit{GWR}$_o$ network after having been trained with the objects from the transitive actions dataset. The first two principal components have been chosen for the visualization in two dimensions.}
\label{fig:obj_clusters}
\end{center}
\end{figure}

A plot showing the neural weights of the~\textit{GWR}$_o$ network is depicted in Fig.~\ref{fig:obj_clusters}. 
Given that the neural weights have a high dimensionality, i.e., the dimensionality of the VLAD descriptors, for illustration purposes we performed principal component analysis (PCA) and show the first two principal components.
As it can be seen from the plot, the neurons are topologically organized into clusters composed of different 2D views of the objects as well as different instances of the same object category. 
This is quite advantageous for our architecture since it allows for generalization towards unseen object views and, to some extent, towards unseen object instances.
The overlap between the \textit{can} and \textit{mug} clusters suggests that the visual appearance of these object categories is more similar than compared to the others and, as a consequence, can be confused.
However, this does not affect the action classification performance, since both of the objects are involved in the same activity, namely \textit{drinking}. 

\begin{figure}
\centering
\includegraphics[width=1.05\linewidth]{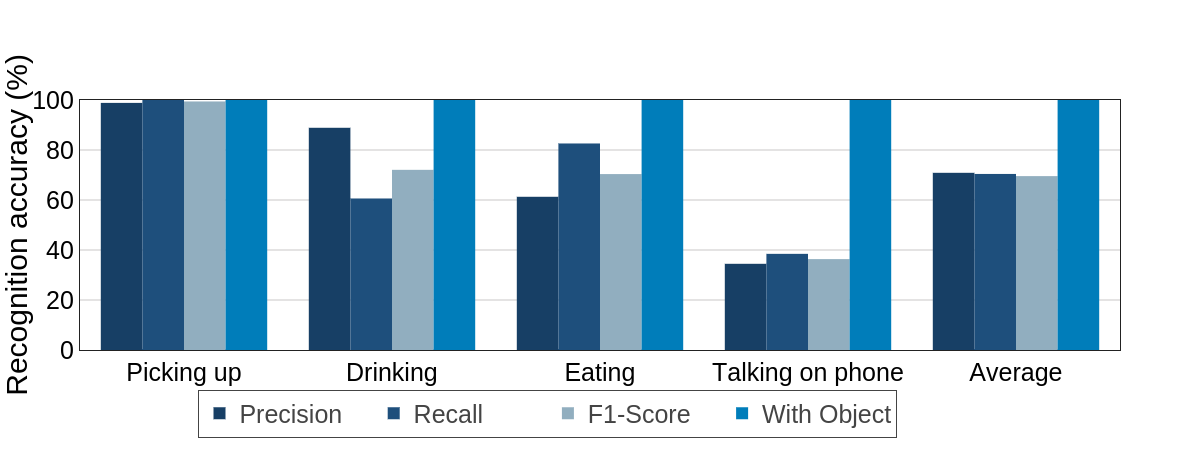}
\caption{Classification results on the transitive actions dataset. Illustrated are precision, recall, and F1-score, averaged over 6 trials of cross-validation when using only the body pose information. When using the manipulated object identity, given by the object recognition module \textit{GWR}$_o$, we obtained a value of $100\%$ for the reported classification performance measures.}
\label{fig:results_mici}
\end{figure}

We report precision, recall, and F1-score~\citep{sokolova2009systematic} for each class of activity, averaged over all six trials in Fig.~\ref{fig:results_mici}.
We obtained values equal to $100\%$ when using the objects' identity information and lower percentage values when using only body pose.
As expected, the increase of the classification performance is more significant for those cases where the body pose introduces ambiguity, e.g., \textit{drinking}, \textit{eating}, \textit{talking on the phone}. 
For the \textit{picking up} activity, on the other hand, the difference in the classification performance is marginal due to the fact that this action can be performed on all of the objects and the identity of a specific object does not play a decisive role.

\subsubsection{Experiments with incongruent action-object pairs}
In addition to the classification experiments, we carried out a qualitative evaluation of the integration module when given in input test data sequences of incongruent action-object pairs.
We consider incongruent pairs to be unusual or functionally irrelevant combinations of actions with objects, e.g. \textit{drinking} with a \textit{telephone} or \textit{eating} with a \textit{can}. 
Interestingly, fMRI studies on human brain have found several regions affected by object-action congruence \citep{yoon2012neural}. 
The neural response in these areas is greater for actions performed on appropriate objects as opposed to unusual actions performed on the same objects.
For this experiment, we artificially created a test dataset, for which we replaced the image of the object being manipulated in each video sequence with the image of an incongruent object extracted from a different action video.

\begin{figure}
\begin{center}
\includegraphics[scale=.11]{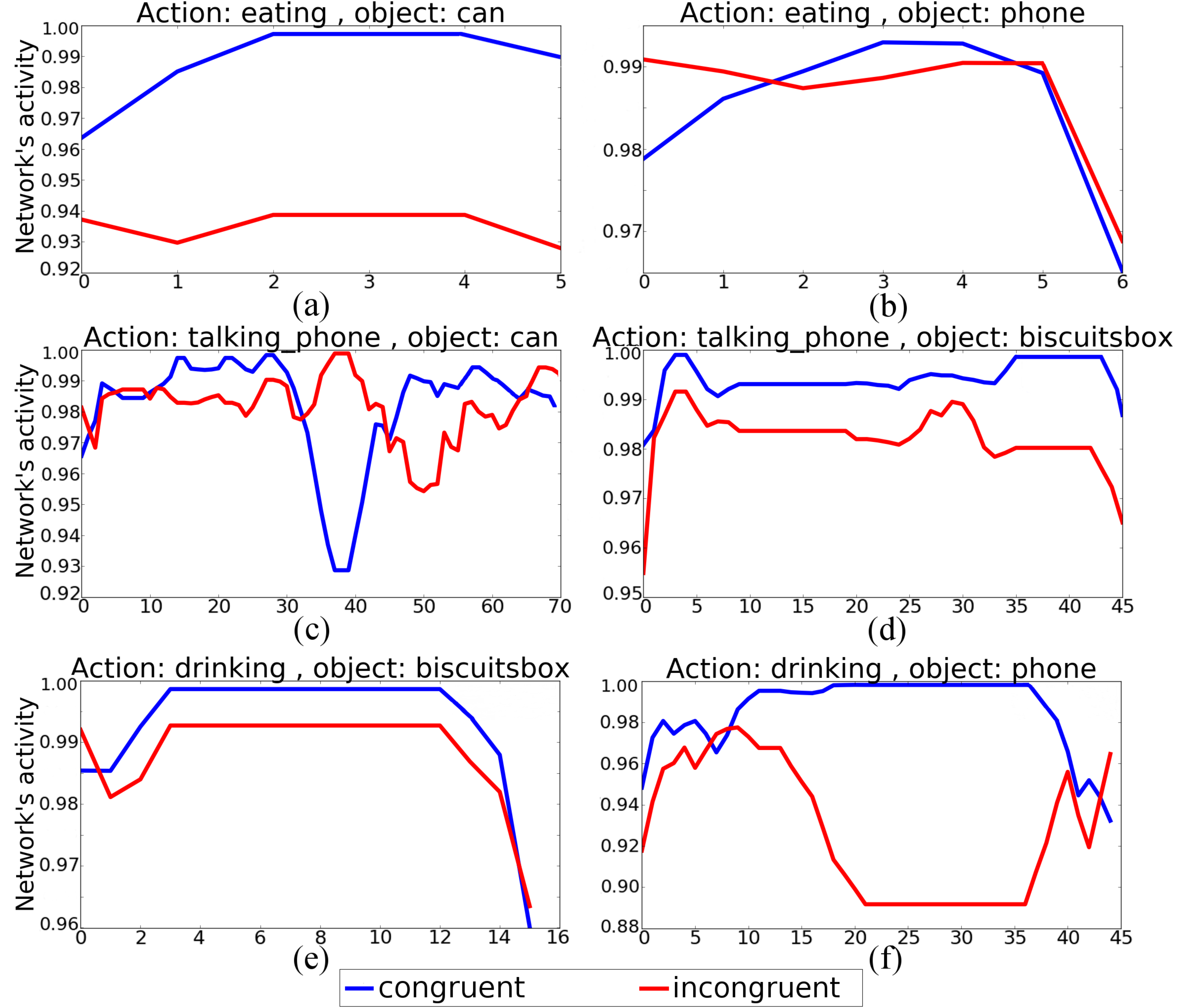}
\end{center}
\caption{Comparison of the \textit{GWR}$_a$ network activations when having as input an action sequence combined with an incongruent object (in red) and one combined with the congruent one (in blue). The $y$ axis represents the activation values, with $1$ being the highest, and the $x$ axis represents the number of frames of the illustrated data sequences. The number of frames can vary among different actions, e.g. the action \textit{eating} is typically shorter than \textit{talking on the phone} and \textit{drinking}.}
\label{fig:plots}
\end{figure}

We analyzed the activation values of the \textit{GWR}$_a$ BMUs (Eq.~\ref{eq:activation}) on both the original action sequence and the manipulated one.
A few examples of the obtained neural activations are illustrated in Fig.~\ref{fig:plots}.
We observed that the activations were typically relatively low for the incongruent samples.
This can be explained by the fact that the \textit{GWR}$_a$ prototypes represent the joint distribution of action segments and congruent objects taken from the congruent set.
The activation of the network is expected to be lower when the input has been taken from a different data distribution than the one the model has learned to fit.
The incongruent samples yield a higher discrepancy with respect to the prototype neurons, thereby leading to a lower network activation.
However, we also noticed some exceptions, e.g., the incongruent pair \textless talking on the phone, can\textgreater ~depicted in Fig.~\ref{fig:plots}.c.
In this case, we can observe that the network activation becomes higher for the incongruent input at a certain point of the sequence, i.e., at a certain action-object segment.
Nevertheless, a decreased network activation on the congruent input indicates that the network has a high quantization error for that particular action-object segment.

It should be noted that a small quantization error of the GWR network is not a requirement for a good performance in the action classification task.
As described in Section~\ref{subsec:classification}, the classification of an action sequence is performed by considering the label histograms associated with the activated neurons. 
We can also notice some cases where the network activation on the incongruent input is not significantly low at the beginning of the sequence, but even slightly higher in the case of \textless eating, phone\textgreater ~(Fig.~\ref{fig:plots}.b).  
A reason for this is the similar motion of the hand holding the object towards the head which may precede both \textit{eating} and \textit{talking on the phone} activities. 
Therefore, exchanging the object \textit{biscuit box} with \textit{phone} for the initial action segments has from little to no impact on the network's response. 

\subsection{Experiments with CAD-120}

\begin{figure}
\begin{center}
\includegraphics[scale=.7]{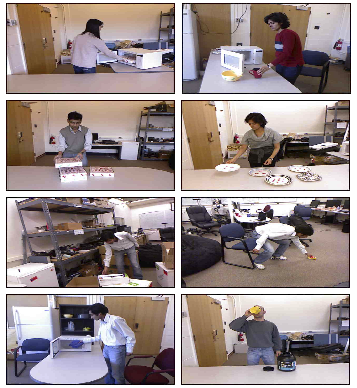}
\caption{Examples of high-level activities from CAD-120 dataset \citep{koppula2013learning}, ordered in each row: \textit{microwaving food}, \textit{taking food}, \textit{stacking objects}, \textit{unstacking objects}, \textit{arranging objects}, \textit{picking objects}, \textit{cleaning objects}, \textit{taking medicine}.}
\label{fig:cad120}
\end{center}
\end{figure}

We evaluated the classification performance of our architecture on the publicly available benchmark dataset CAD-120 (Fig.~\ref{fig:cad120}).
This dataset consists of 120 RGB-D videos of 10 long daily activities: \textit{arranging objects}, \textit{cleaning objects}, \textit{having meal}, \textit{making cereal}, \textit{microwaving food}, \textit{picking objects}, \textit{stacking objects}, \textit{taking food}, \textit{taking medicine} and \textit{unstacking objects}. These activities are performed by four different subjects (two males, two females and, of these four, one left-handed) repeating each action three to four times.  
Each video is annotated with the human skeleton tracks and the position of the manipulated objects across frames. 

We computed skeletal quad features (described in Section. \ref{subsec:data_proc}) for the encoding of the pose of the upper body, based on the three-dimensional position of skeletal joints provided in the dataset.
Additionally, we extracted RGB images of manipulated objects from each frame and encoded them through VLAD encoding technique as described in Section.~\ref{subsec:data_proc}.
For the concatenation of the processed body pose cues, a time window of nine frames was chosen.
Since we down-sample the activity video frames to a rate of $10$ fps, this leads to an action-object segment having a temporal duration of $0.9$ seconds.
After training the whole architecture with an input data of $\approx18.000$ frames, the number of neurons reached in each GWR network was $460$ for \textit{GWR}$_b$ , $410$ for \textit{GWR}$_o$, while for \textit{GWR}$_a$ the number varied from $\approx3200$ to $\approx3700$ across different trials of the cross-validation.

\begin{figure}
\begin{center}
\includegraphics[scale=.42]{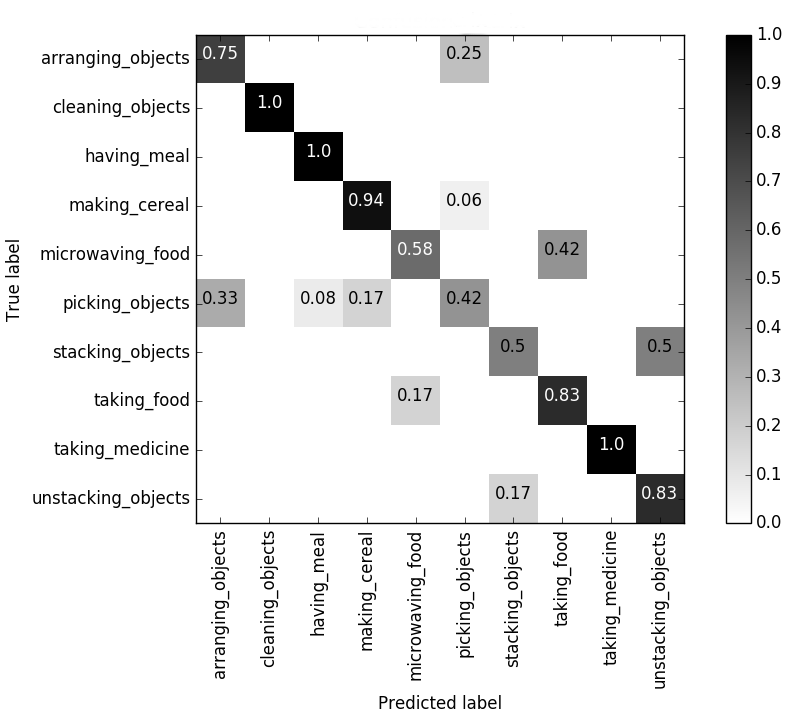}
\end{center}
\caption{Confusion matrix for the 10 high-level activities of CAD-120 dataset.}
\label{fig:conf_cad}
\end{figure}

In Fig.~\ref{fig:conf_cad}, we show the confusion matrix for the 10 high-level activities of this dataset. 
We inspected that the activities interchanged by our model were the ones including the same category of objects and similar body motions, e.g., \textit{stacking objects} and \textit{unstacking  objects}, \textit{microwaving food} and \textit{taking food}. 
Also, the activity of \textit{picking objects} was often confused with \textit{arranging objects}, due to the fact that body pose segments of the first are similar to the ones preceding the activity of \textit{arranging objects}.
In Table \ref{table:cad120}, we show a comparison of our results with the state of the art on the CAD-120 dataset with accuracy, precision, and recall as evaluation metrics.
We obtained 79\% of accuracy, 80.5\% of precision, and 78.5\% of recall.

We reported only the approaches that do not use ground-truth temporal segmentation of the activities into smaller atomic actions or sub-activities~\cite{hu2014learning, taha2015skeleton}.
Our results are comparable with Rybok et al.~\cite{rybok2014important}.
Similar to our work, their method considers objects' appearance as contextual information which is then concatenated with body motion features represented as a bag of words. 
The best results were obtained by Koppula et al.~\cite{koppula2013l}, reporting 83.1\% of accuracy, 87\% of precision and 82.7\% of recall.
In their work, spatiotemporal dependencies between actions and objects are modelled by a Conditional Random Field (CRF) which combines and learns the relationship between a number of different features such as the coordinates of the object's centroid, the total displacement and the total distance moved by the object's centroid in each temporal segment, the difference in $(x, y, z)$ coordinates of the object and skeleton joint locations and their distances.
After the generation of the graph which models spatiotemporal relations, they use a Support Vector Machine (SVM) for classifying action sequences. 
Unlike our work, they do not perform object classification but rely on manually annotated labels.

\begin{table}
\centering
\resizebox{.75\columnwidth}{!}{%
  \begin{tabular}{ l|c|c|c|c|c|c }
    \hline
    \textbf{Algorithm} & \textbf{U} & \textbf{O. Rec.} & \textbf{O. Tr.} & \textbf{Acc.} & \textbf{Prec.} & \textbf{Rec.} \\ \hline
	\begin{tabular}{@{}c@{}}Koppula et al.\citep{koppula2013l}, \\ (\textit{CRF},~\textit{SVM})\end{tabular}  & - & - & \checkmark & 83.1 & 87.0 & 82.7 \\ \hline
    \begin{tabular}{@{}c@{}}Koppula et al.\citep{koppula2013learning}, \\ (\textit{CRF},~\textit{SVM})\end{tabular} & - & - & \checkmark & 80.6 & 81.8 & 80.0 \\ \hline 
    \begin{tabular}{@{}c@{}}\textbf{Our approach}, \\ \textbf{(GWR)} \end{tabular} & \checkmark & \checkmark & - & \textbf{79.0} & \textbf{80.5} & \textbf{78.5} \\   \hline   
    \begin{tabular}{@{}c@{}}Rybok et al.\citep{rybok2014important}, \\ (\textit{SVM})\end{tabular} & - & \checkmark & - & 78.2 & - & - \\ \hline
    \begin{tabular}{@{}c@{}}Tayyub et al.\citep{Tayyub}, \\ (\textit{SVM})\end{tabular} & - & - & \checkmark & 75.8 & 77.9 & 75.4 \\ \hline
  \end{tabular}
  }
    \caption[Table caption text]{Results on the CAD-120 dataset for the recognition of 10 high-level activities. Reported are accuracy, precision and recall (in percentage) averaged over 4-fold cross-validation experiments. For comparison, we have included which of the reported methods is unsupervised (U), performs object recognition for the classification of the activities (O.Rec.) or relies on object tracking (O.Tr.).}
    \label{table:cad120}
\end{table}

We assume that the tracking of the objects' position in the scene as well as the objects' distance from the subject's hand provides additional information that might improve our classification results and is considered part of our future work.
 
\section{Discussion}\label{sec:discussion}

\subsection{Summary}

In this paper, we presented a self-organizing neural network architecture that learns to recognize actions comprising human-object interaction from RGB-D videos.
Our architecture consists of two pathways of GWR networks processing respectively body pose and object appearance and identity, with a subsequent integration layer learning action-object mappings.
The prototype-based learning mechanism of the GWR allows to attenuate input noise and to generalize towards unseen data samples.
For the purpose of classification, we extended the GWR with a labeling technique based on majority vote. 

The evaluation of our approach has shown good results on a dataset of human-object interactions collected specifically for the study of the importance of the identity of objects.
The analysis of the neural response of the integration layer showed an overall lower network activation when given incongruent action-object pairs compared to the congruent pairs.
Furthermore, the classification accuracy of our unsupervised architecture on a publicly available action benchmark dataset is competitive with respect to the supervised state-of-the-art approaches. 

\subsection{Self-organizing neural learning and analogies with neuroscience}

Generative approaches based on self-organization learn input probability distribution through a finite set of reference vectors associated with neurons.
Moreover, they resemble the topological relationships of the input space through the neurons' organization. 
Growing self-organizing approaches such as the GNG \citep{fritzke1995growing} and the GWR networks \citep{marsland2002self} are characterized by a dynamic topological structure able to adapt toward the input data space through the mechanism of the competitive Hebbian learning \citep{martinetz1993competitive}. 
Unlike the GNG, where the network grows at a constant rate, the GWR algorithm is equipped with a learning mechanism that creates new neurons whenever the current input is not well represented by the prototype neurons.

We extended the GWR algorithm, which processes input data vectors in the spatial domain, to the processing of temporal data by the mechanism of the temporal sliding window \citep{parisi2014human}.
The temporally ordered neural activations obtained through this technique resemble the motion pattern encodings through the snapshot neurons found in the STS area of the brain \citep{giese2003neural}. 
From the computational perspective, the sliding window technique allows for the extrapolation of spatiotemporal dependencies in the data sequences. 
The use of prototype-based representations for objects is motivated by psychological studies on the nature of human categorization \citep{rosch1975family}. 
According to the exemplar-based theory, categories of objects and concepts are typically learned as a set of prototypical examples and the similarity, or the so-called family resemblance, is used for class association. 

Finally, the use of the GWR for integrating information about action and objects produced a behavior resembling the action-selective neural circuits which show sensitivity to the congruence of the action being performed on an object \citep{yoon2012neural}.  

\subsection{Future Work}

In this work, we focused on a two-pathway hierarchy for learning human-object interactions represented as a combination of upper body pose configurations and objects' category labels. 
However, in order to reduce the computational complexity of the architecture, we have excluded an important component: the motion information.
Results from other approaches on recognition of human-object interactions and on the learning of object affordances \cite{kjellstrom2011visual, koppula2013learning} have shown that tracking the object's position and spatial relationship with respect to the body can help for a better interpretation of this type of interaction. 
There is evidence from neuroscience that the observation of using a tool activates areas of the lateral temporal cortex in the human brain which is engaged in perceiving and storing information about motion \citep{beauchamp2002parallel}. 
Neural mechanisms for the processing of human body motion are also believed to contribute to action discrimination in general \citep{giese2003neural}. 
Therefore, a possible next step is to extend our model by including motion information.
 
An additional future work direction is the introduction of recurrent connections in the GWR networks for the purpose of temporal sequence processing. Recurrence in self-organizing networks has been extensively investigated and applied to temporal data classification \citep{strickert2005merge,PW17}.
In the current implementation, temporal dependencies are encoded and learned by hard assignments to time windows. 
However, the concatenation of perceptual feature vectors may lead to very high-dimensional spaces, whereby methods based on a Euclidean distance metric are known to perform worse~\citep{aggarwal2001surprising}. 

In our current work, we used depth information for the efficient extraction of a three-dimensional skeleton model.
However, when dealing with more complex activities such as human-object interactions, this type of depth representation may be subject to a number of issues such as a highly noisy skeleton due to body self-occlusions when manipulating an object.
Therefore, future work will address the limitations of this hand-crafted feature extraction with a neural architecture able to extract visual features from raw images, e.g., with the use of deep neural network self-organization~\citep{PW17}.

Finally, the results reported in this paper motivate future work towards the integration of our learning system into a robotic platform and its evaluation in real-world scenarios such as learning by imitation tasks or human-robot assistance in natural environments. 

\section*{Acknowledgments}
The authors gratefully acknowledge partial support by the EU- and City of Hamburg-funded program Pro-Exzellenzia 4.0, the German Research Foundation DFG under project CML (TRR 169), and the Hamburg Landesforschungsf{\"o}rderungsprojekt.

\section*{References}

\bibliography{biblio}

\end{document}